\pdfoutput=1
\documentclass[11pt]{article}

\usepackage[]{EMNLP2023}

\usepackage{times}
\usepackage{latexsym}

\usepackage[T1]{fontenc}
\usepackage[utf8]{inputenc}

\usepackage{microtype}

\usepackage{inconsolata}

\usepackage{CJKutf8} % 中文
\usepackage{graphicx}
\usepackage{caption}
\usepackage{float}
\usepackage{algorithmic}
\usepackage{algorithm}
\usepackage{array}
\usepackage[caption=false,font=normalsize,labelfont=sf,textfont=sf]{subfig}
\usepackage{multirow}
\usepackage{booktabs}
\usepackage{makecell}
\usepackage{colortbl} % https://www.jianshu.com/p/c6c5655b9b7c

\title{SoulChat: Improving LLMs' Empathy, Listening, and Comfort Abilities through Fine-tuning with Multi-turn Empathy Conversations}

\author{
    Yirong Chen\textsuperscript{1}, Xiaofen Xing\textsuperscript{1}\Thanks{ Corresponding author. Email: xfxing@scut.edu.cn}, Jingkai Lin\textsuperscript{1}, Huimin Zheng\textsuperscript{1}, \\
    {\bf Zhenyu Wang\textsuperscript{1}, Qi Liu\textsuperscript{2}, Xiangmin Xu\textsuperscript{2,3}} \\
  \textsuperscript{1}Guangdong Provincial Key Laboratory of Human Digital Twin, School of EE., \\ South China University of Technology, Guangzhou, China \\
  \textsuperscript{2}School of Future Technology, South China University of Technology, Guangzhou, China \\
  \textsuperscript{3}Pazhou Lab, Guangzhou, China \\
  \texttt{eeyirongchen@mail.scut.edu.cn}, 
  \texttt{\{xfxing, xmxu\}@scut.edu.cn} \\
 }

\begin{document}
\maketitle
\begin{abstract}
Large language models (LLMs) have been widely applied in various fields due to their excellent capability for memorizing knowledge and chain of thought (CoT). When these language models are applied in the field of psychological counseling, they often rush to provide universal advice. However, when users seek psychological support, they need to gain empathy, trust, understanding and comfort, rather than just reasonable advice. To this end, we constructed a multi-turn empathetic conversation dataset of more than 2 million samples, in which the input is the multi-turn conversation context, and the target is empathetic responses that cover expressions such as questioning, comfort, recognition, listening, trust, emotional support, etc. Experiments have shown that the empathy ability of LLMs can be significantly enhanced when finetuning by using multi-turn dialogue history and responses that are closer to the expression of a psychological consultant.\footnote{\url{https://github.com/scutcyr/SoulChat}}
\end{abstract}

\section{Introduction}
% 第1段：描述大模型的进展
With the birth of BERT~\citep{devlin-etal-2019-bert} and GPT~\citep{radford2018improving}, large language models (LLMs) have made rapid progress in the past five years. In November 2022, OpenAI launched ChatGPT\footnote{\url{https://chat.openai.com}}~\citep{chatgpt}, a large language model fine-tuning by reinforcement learning from human feedback (RLHF)~\citep{instructgpt}. However, when applied to mental health or emotional support conversation, there are three main issues lead to ChatGPT appear less ``human-centered'':
\begin{itemize}
  \item [1)]
  \textbf{ChatGPT tends to provide repetitive and standardized responses.} ChatGPT often uses the following template to respond to users' questions related to mental health: "\begin{CJK}{UTF8}{gbsn}我很抱歉...。xxx是...。以下是一些建议：...。\end{CJK} (I'm sorry to ...\{xxx\} is ...Here are some suggestions:...)", which may cause boredom.
  \item [2)]
  \textbf{ChatGPT is inclined to provide suggestions rather than ask questions or listen.} It is eager to solve users' problems, usually providing lengthy and general suggestions, as shown in Figure~\ref{ChatGPT_Example} of Appendix~\ref{sec:appendix_sample}. However, professional psychologists rarely provide specific suggestions during the counseling process.
  \item [3)]
  \textbf{ChatGPT acts a bit like a rational "Straight man" for those users who need listening and comfort.}  Users who seek emotional support usually expect empathy support such as listening, understanding and comfort.
\end{itemize}

\begin{figure*}[ht]
  \centering
  \includegraphics[width=\textwidth]{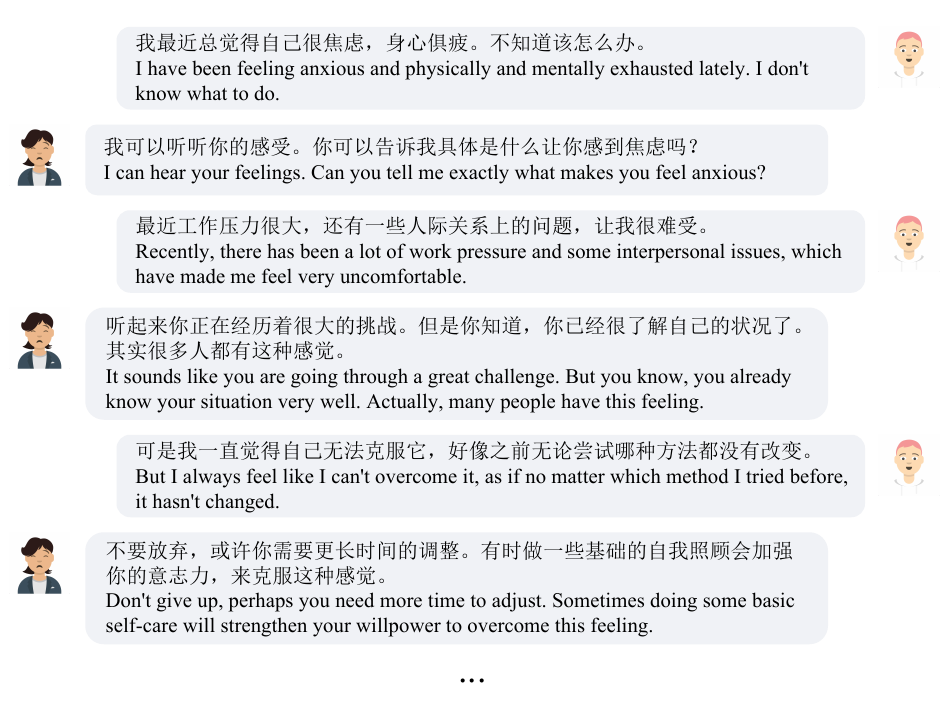}
  \caption{A case of a user confiding to SoulChat. Compared to ChatGPT, SoulChat is better at listening and guiding users to think.
}
  \label{SoulChat_Example}
\end{figure*}

% \footnote{\url{https://chatglm.cn}}
Similar issues can also be found in other LLMs, e.g. ChatGLM~\citep{zeng2023glm-130b}, SparkDesk\footnote{\url{https://xinghuo.xfyun.cn}}, as presented in Appendix~\ref{sec:appendix_sample}. It may be due to the lack of large-scale multi-turn empathy conversation datasets for fine-tuning stage, especially in the field of Chinese mental health or emotional support. EMPATHETICDIALOGUES~\citep{rashkin-etal-2019-towards} and ESConv~\citep{liu-etal-2021-towards} are two English empathy conversation datasets that is used for developing emotional support conversation (ESC) systems, e.g MISC~\citep{tu-etal-2022-misc}, GLHG~\citep{peng2022control}, MultiESC~\citep{cheng-etal-2022-improving}, FADO~\citep{PENG2023110340} and etc. On the one hand, these models may rely on annotated empathy strategies and emotions of users during the training or inference stage, which means that building large-scale similar datasets for fine-tuning LLMs is difficult. On the other hand, these datasets are in English, so that they cannot be applied to fine-tune Chinese LLMs. As for mental health, efaqa~\citep{efaqa-corpus-zh:petpsychology} and PsyQA~\citep{sun-etal-2021-psyqa} are two commonly-used datasets. Among them, efaqa contains 20,000 conversations and provides annotation information such as types of troubles, psychological disorders, SOS, etc. However, efaqa has a complex multi-party dialogue relationship and a high proportion of low-quality responses from netizens, while PsyQA contains 22,346 questions and 56,063 single-turn long-text psychological counseling conversations. Thus, neither of these datasets can solve the three issues of ChatGPT mentioned above. 

Recently, \citet{qiu2023smile} proposed a SMILE approach to employ ChatGPT to convert single-turn dialogues into multi-turn ones. They utilized SMILE to extend the single-turn conversation dataset PsyQA to a empathy multi-turn conversation dataset SMILECHAT with 355,733 samples. Inspired by \citep{qiu2023smile}, we proposed a Chinese empathy constraint prompt, in which the empathy prompt constraint is further strengthened compared with SMILE prompt (see Appendix~\ref{sec:prompt_analysis}). As shown in Figure~\ref{prompt} (English version: Appendix~\ref{sec:prompt_analysis}), our empathy constraints are defined as ``\begin{CJK}{UTF8}{gbsn}`心理咨询师'的回复需要结合用户的描述内容并提供共情，如：倾听、安慰、理解、信任、认可、真诚、情感支持等\end{CJK} (The response of the 'psychological counselor' needs to be combined with the user's description and provide empathy, such as listening, comfort, interpretation, trust, recognition, sincerity, emotional support, etc)''.

\begin{figure}[htbp]
  \centering
  \includegraphics[width=0.48\textwidth]{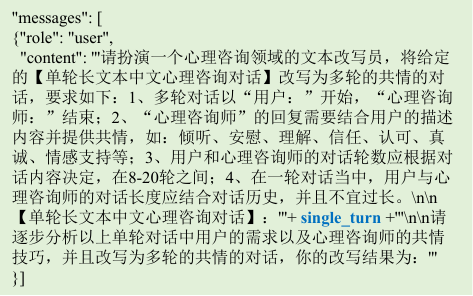}
  \caption{The prompt used for converting single-turn psychological counseling conversations to multi-turn empathy conversations (English version: Appendix~\ref{sec:prompt_analysis}). 
  %The variable  "single\_turn" represents a single-turn conversation. "$\backslash$n" is a line break. "+" indicates string concatenation.
}
  \label{prompt}
\end{figure}

To this end, we first constructed 215,813 different psychological counseling questions about 12 topics and 619,725 answers through data outsourcing services. Rule-based cleaning, manual rewriting and human proofreading are applied to ensure that there is no sensitive or privacy-related content in the dataset. Then, we use ChatGPT to convert these single-turn long text psychological counseling conversations to multi-turn empathy conversations. We also conducted manual proofreading and data cleansing for multi-turn dialogues rewritten by ChatGPT to further strengthen the expression of empathy, such as questioning, comfort, recognition, listening, trust, emotional support, etc. In the end, we obtained a  multi-turn empathy conversation dataset, named SoulChatCorpus, with 2,300,248 samples. To our knowledge, it is the first million-scale multi-turn empathy conversation dataset in the field of mental health or emotional support. We conduct experiments by using ChatGLM-6B as the base model for fine-tuning on SoulChatCorpus. Results demonstrate that LLMs' empathy, listening, and comfort abilities can be improved significantly through fine-tuning with million-scale multi-turn empathy conversation dataset.

%\section{Related Work}
%\label{sec:related_work}
%\subsection{Large Language Model}
%Large language models (LLMs) has achieved excellent performance in various fields in the past five years.
%
%\subsection{Mental Health Dataset}

\begin{figure}[htbp]
  \centering
  \includegraphics[width=0.5\textwidth]{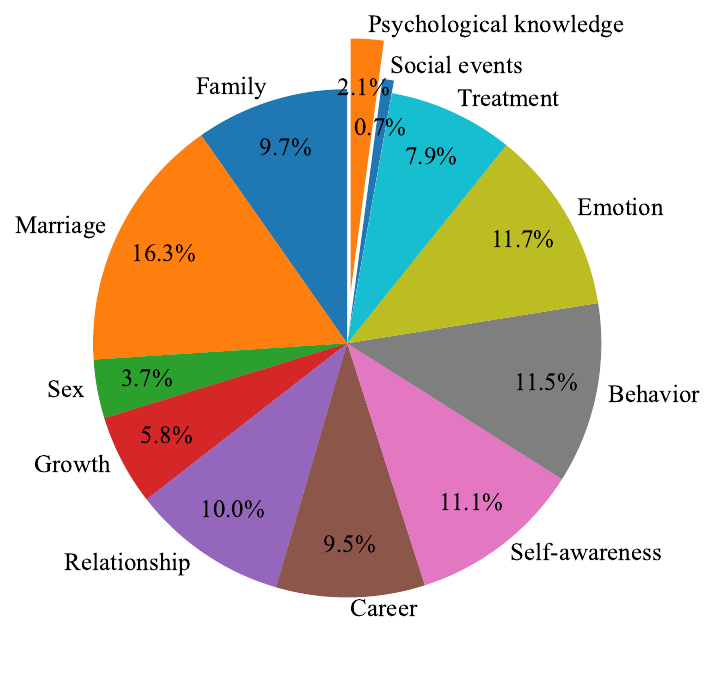}
  \caption{Distribution of counseling topics.
}
  \label{tag_ratio}
\end{figure}

\section{Human-centered Mental Health LLM}
\label{sec:soulchat}
\subsection{SoulChatCorpus Collection}
\begin{figure}[htbp]
  \centering
  \includegraphics[width=0.5\textwidth]{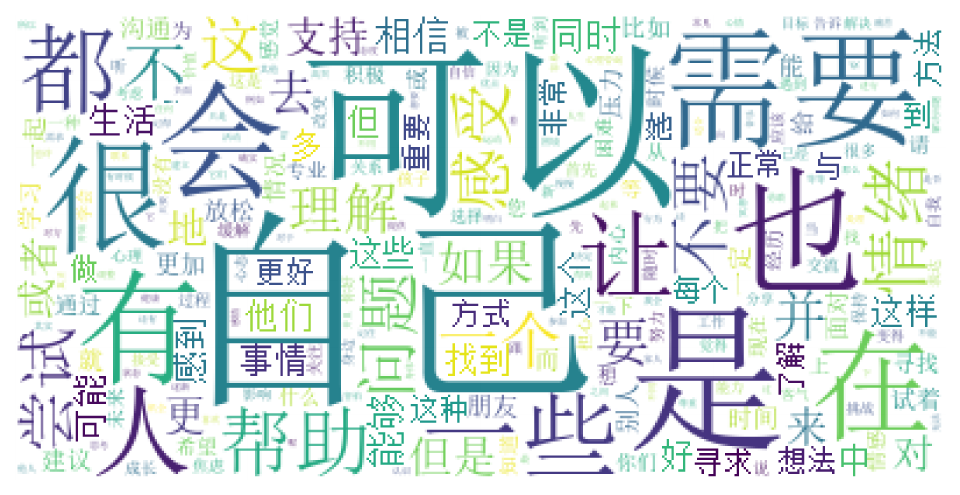}
  \caption{Word cloud map of psychological consultants' utterances (English version: Appendix~\ref{sec:english_target_words}).
}
  \label{target_words}
\end{figure}

We consider an one-on-one psychological counseling conversational setting where a user and a psychological consultant engage in multiple rounds of dialogue. However, such conversation data is not publicly available due to the privacy protection and ethical standards of psychological counseling. To construct high-quality multi-turn empathy conversation dataset, We selected 12 topics of psychological counseling to construct 215,813 long-text questions and 619,725 long-text answer through crowdsourcing. The distribution of topics is shown in Figure \ref{tag_ratio}. Then, we used ChatGPT (99\% called gpt-3.5-turbo api and 1\% called gpt-4 api) as a text rewriting tool following the prompt as shown in Figure \ref{prompt} to convert single-turn psychological counseling conversations to multi-turn empathy conversations, in which one turn is in the form of "\begin{CJK}{UTF8}{gbsn}用户：\end{CJK} <user\_utt>$\backslash$n\begin{CJK}{UTF8}{gbsn}心理咨询师：\end{CJK}<psy\_utt>". The response of "\begin{CJK}{UTF8}{gbsn}心理咨询师\end{CJK}" was asked to be rewritten to reflect human-centered expressions such as empathy, listening, comfort, etc. Finally, after manual proofreading, we removed 105,134 low-quality samples and ultimately obtained 2,300,248 samples. As shown in Figure~\ref{target_words}, the word cloud map of the utterances expressed by psychological consultants indicated that the rewritten multi-turn empathy conversation has high level of empathy.

\begin{table*}[htbp]
\caption{\label{evaluation_results} Evaluation results.}
\centering
\begin{tabular}%{clm{0.7cm}m{0.7cm}m{0.7cm}m{0.7cm}m{0.7cm}m{0.7cm}m{0.8cm}m{0.7cm}m{0.7cm}m{0.7cm}}
{clccccccccccc}
\toprule
% 表头
\multirow{2}{*}{Dataset} & \multirow{2}{*}{Model} & \multicolumn{7}{c}{Automatic.} & \multicolumn{3}{c}{Manual.}
    \\\cmidrule(lr){3-9}\cmidrule(lr){10-13}
           &  & \small{B-1} & \small{B-2} & \small{B-3} & \small{B-4} & \small{R-1} & \small{R-2} & \small{R-L} & \small{Con.} & \small{Emp.} & \small{Hel.} & \small{Saf.}\\\midrule

\multirow{4}{*}{\small{\shortstack{\small{SoulChat-} \\\small{Corpus}}}} 
& \small{ChatGLM-6B} & \small{22.73} & \small{13.15} & \small{8.04} & \small{4.92} & \small{25.33} & \small{5.72} & \small{18.84} & \small{1.90} & \cellcolor{green!20} \small{1.55} & \small{1.92} & \small{1.0}\\
& \small{MeChat} & \small{29.43} & \small{17.12} & \small{10.54} & \small{6.71} & \small{27.35} & \small{6.27} & \small{21.12} & \small{1.83} & \cellcolor{green!20} \small{1.70} & \small{1.78} & \small{1.0}\\
& \small{ChatGPT} & \small{27.98} & \small{16.09} & \small{9.93} & \small{6.23} & \small{27.39} & \small{6.82} & \small{21.92} & \small{\textbf{1.96}} & \cellcolor{green!20} \small{1.62} & \small{\textbf{1.94}} & \small{1.0}\\
& \small{SoulChat} & \small{\textbf{33.78}} & \small{\textbf{20.07}} & \small{\textbf{12.86}} & \small{\textbf{8.52}} & \small{\textbf{31.47}} & \small{\textbf{8.92}} & \small{\textbf{26.57}} & \small{1.95} & \cellcolor{green!20} \small{\textbf{1.84}} & \small{1.87} & \small{1.0}\\
\midrule
\multirow{4}{*}{\small{SMILECHAT}} 
& \small{ChatGLM-6B} & \small{22.91} & \small{13.56} & \small{8.40} & \small{5.15} & \small{25.99} & \small{5.95} & \small{18.76} & \small{1.81} & \cellcolor{green!20} \small{1.39} & \small{1.84} & \small{1.0}\\
& \small{MeChat} & \small{30.63} & \small{18.41} & \small{11.59} & \small{7.46} & \small{28.92} & \small{6.76} & \small{21.59} & \small{\textbf{1.95}} & \cellcolor{green!20} \small{1.74} & \small{1.83} & \small{1.0}\\
& \small{ChatGPT} & \small{28.30} & \small{16.48} & \small{10.24} & \small{6.40} & \small{27.57} & \small{6.71} & \small{\textbf{21.60}} & \small{\textbf{1.95}} & \cellcolor{green!20} \small{1.65} & \small{\textbf{1.97}} & \small{1.0}\\
& \small{SoulChat} & \small{\textbf{35.40}} & \small{\textbf{21.39}} & \small{\textbf{13.77}} & \small{\textbf{9.02}} & \small{\textbf{32.64}} & \small{\textbf{9.17}} & \small{21.10} & \small{1.93} & \cellcolor{green!20} \small{\textbf{1.90}} & \small{1.85} & \small{1.0}\\
\bottomrule
\end{tabular}
\end{table*}

\subsection{SoulChat Model}
We utilized the ChatGLM-6B~\citep{du-etal-2022-glm, zeng2023glm-130b} as the base LLM architecture to develop the SoulChat. ChatGLM-6B is an open-source, bilingual LLM based on the General Language Model (GLM)~\citep{du-etal-2022-glm} framework with 6.2 billion parameters. The input of model is defined as: 
$$
input = u_1^u+'{\backslash}n'+u_1^p+ ...+u_N^u+'{\backslash}n'+u_N^p
$$
where the utterance of User on $i$ turn $u_i^u$=`\begin{CJK}{UTF8}{gbsn}用户：\end{CJK} (User:)' + utterance$_{i}^u$,  utterance of Psychologist on $i$ turn $u_i^p$=`\begin{CJK}{UTF8}{gbsn}心理咨询师：\end{CJK} (Psychologist:)' + utterance$_{i}^p$ ($i<N$), $u_N^p$=`\begin{CJK}{UTF8}{gbsn}心理咨询师：\end{CJK} (Psychologist:)', $N$
represents the number of conversation turns for the context.

\section{Experiments}
\label{sec:experiments}
\subsection{Baselines}
We compare SoulChat and the following benchmark models using both automatic and manual evaluations:
\begin{itemize}
  \item [1)]
  \textbf{ChatGLM-6B}\footnote{\url{https://github.com/THUDM/ChatGLM-6B}}~\citep{du-etal-2022-glm, zeng2023glm-130b} serves as the base model for SoulChat.
  \item [2)]
  \textbf{ChatGPT}~\citep{chatgpt, instructgpt} is a LLM that is trained using supervised finetuning and Reinforcement Learning from Human Feedback (RLHF).
  \item [3)]
  \textbf{MeChat}~\citep{qiu2023smile} is a LLM finetuned with low-rank adaptation (LoRA)~\citep{hu2022lora} on SMILECHAT dataset that is generated by ChatGPT based on PsyQA.
  
\end{itemize}

\subsection{Implementation details}
SoulChat is finetuned on the proposed SoulChatCorpus with a batch size of 80 and global training steps of 30,000. The \textit{WarmupDecayLR} learning rate scheduler with $warmup\_steps = 1000$ and $warmup\_max\_lr=5e-5$ is used. The maximum input token length is set to 1,536. The maximum target token length is set to 512. The decoding algorithms of Top-p sampling with $p=0.75$ and temperature $\tau=0.95$ is adopted during the inference phase.

\subsection{Results and Analysis}
We randomly selected 10,000 samples from SoulChatCorpus and SMILECHAT respectively as the test set for automatic evaluation and 100 samples for manual evaluation. For each sample, each model generates an answer for evaluation. We used 7 evaluation metrics as automatic metrics: BLEU-1 (B-1), BLEU-2 (B-2), BLEU-3 (B-3), BLEU-4 (B-4)~\citep{papineni-etal-2002-bleu}, R-1 (ROUGE-1), R-2 (ROUGE-2) and R-L (ROUGE-L)~\citep{lin-2004-rouge}). Three individual experts majoring in \textit{Psychology} were asked to evaluate the generated responses in terms of content naturalness (Con.), empathy level (Emp.), Helpfulness (Hel.) and Safety (Saf.), as detailed described in Appendix~\ref{sec:manual_evaluation_instructions}. The rating scale of Con., Emp. and Hel. is $(0, 1, 2)$, while $(0, 1)$ for Saf., where higher score means better. One hundred dialogues were randomly sampled from the test set of SoulChatCorpus and SMILECHAT for manual evaluation. Fleiss' $\kappa$ \cite{fleiss1971measuring} for Con., Emp. and Hel. are 0.489, 0.472 and 0.532, indicating moderate annotation agreement respectively, while $\kappa=1$ for Saf. (perfect agreement). The evaluation results are shown in Table~\ref{evaluation_results}. Generally, SoulChat outperforms ChatGLM-6B, ChatGPT and MeChat in both automatic evaluation metrics and Emp. metric on test set of SoulChatCorpus and SMILECHAT. Specifically, the results on SMILECHAT demonstrates SoulChat's excellent zero-shot performance in the field of mental health.

\section{Conclusion and Future Work}
\label{sec:conclusion}
In this paper, we explore how to make LLMs more human-centered. To this end, we constructed a Chinese large-scale multi-turn empathy conversation dataset, named SoulChatCorpus, with 12 empathy topics and more than 2 million samples. The experimental results indicate that using this dataset to finetune LLMs leads to high-level empathy ability when users try to seek emotional support from LLMs. Future work needs to further consider user attributes, such as personality, gender and etc., to help LLMs generate targeted empathy responses for different individuals.

\section*{Limitations}
In this work we proposed a human-centered LLM named SoulChat that has excellent empathy ability, which is finetuned on the proposed SoulChatCorpus dataset. Although the experimental results demonstrate the effectiveness of SoulChat, there are still some limitations need to consider. The mechanism of empathy is complex. Different users have different expectations for the output of the model. For example, when discussing tense emotions, there are significant differences in the solutions expected by adults and adolescents. Therefore, human-centered LLMs need to further consider the user's personality, identity, and other attributes to assist in generating answers that are closer to the user's needs.

\section*{Ethics Statement}
\begin{itemize}
  \item
  \textbf{Data Collection.} In order to protect privacy~\citep{hovy-spruit-2016-social}, we adopted strict manual proofreading process when constructing the dataset. We filtered all samples with special strings such as "\begin{CJK}{UTF8}{gbsn}我是\end{CJK} (I am)", "\begin{CJK}{UTF8}{gbsn}自杀\end{CJK} (suicide)", "\begin{CJK}{UTF8}{gbsn}跳楼\end{CJK} (jumping off a building)", etc., and conducted manual data cleansing. Any text related to privacy has been rewritten or removed. Besides, any potential conversations that pose harm to users, others, or society have been completely removed from our data. To this end, we removed 105,134 samples from multi-turn conversations generated by ChatGPT.
  \item
  \textbf{Potential Risks of the Model} We conducted a safety assessment specifically for the output of the model during the manual evaluation phase, and the results are shown in Table~\ref{evaluation_results}. Due to the lack of human feedback during the model finetuning stage, there are inevitably answers that may pose harm to users. Therefore, future work needs to combine RLHF to improve the safety level of model generated content. In addition, when this model is applied to downstream scenarios, it is necessary to inform the users in advance that the answers they see are generated by the AI model and are for reference only.
  \item
  \textbf{Annotator Compensation.} We invited individual experts majoring in \textit{Psychology} to conduct the proposed \textbf{CEHS} evaluation of the model's output. The annotators' evaluation of each sample takes approximately 3 minutes, during which they can receive a salary of \$0.418. Therefore, the hourly salary of the annotators is \$8.36, which is higher than the US minimum wage of \$7.12 per hour.
\end{itemize}

\section*{Acknowledgements}
This work was supported by the Science and Technology Project of Guangzhou (202103010002), the Natural Science Foundation of Guangdong Province (2022A1515011588), the National Key R\&D Program of China (2022YFB4500600), the Science and Technology Project of Guangdong (2022B0101010003), the National Natural Science Foundation of China under Grant U1801262 and Guangdong Provincial Key Laboratory of  Human Digital Twin (2022B1212010004).


\begin{thebibliography}{20}
\expandafter\ifx\csname natexlab\endcsname\relax\def\natexlab#1{#1}\fi

\bibitem[{Cheng et~al.(2022)Cheng, Liu, Li, Wang, Zhao, Liu, Liang, and Zheng}]{cheng-etal-2022-improving}
Yi~Cheng, Wenge Liu, Wenjie Li, Jiashuo Wang, Ruihui Zhao, Bang Liu, Xiaodan Liang, and Yefeng Zheng. 2022.
\newblock \href {https://aclanthology.org/2022.emnlp-main.195} {Improving multi-turn emotional support dialogue generation with lookahead strategy planning}.
\newblock In \emph{Proceedings of the 2022 Conference on Empirical Methods in Natural Language Processing}, pages 3014--3026, Abu Dhabi, United Arab Emirates. Association for Computational Linguistics.

\bibitem[{Devlin et~al.(2019)Devlin, Chang, Lee, and Toutanova}]{devlin-etal-2019-bert}
Jacob Devlin, Ming-Wei Chang, Kenton Lee, and Kristina Toutanova. 2019.
\newblock \href {https://doi.org/10.18653/v1/N19-1423} {{BERT}: Pre-training of deep bidirectional transformers for language understanding}.
\newblock In \emph{Proceedings of the 2019 Conference of the North {A}merican Chapter of the Association for Computational Linguistics: Human Language Technologies, Volume 1 (Long and Short Papers)}, pages 4171--4186, Minneapolis, Minnesota. Association for Computational Linguistics.

\bibitem[{Du et~al.(2022)Du, Qian, Liu, Ding, Qiu, Yang, and Tang}]{du-etal-2022-glm}
Zhengxiao Du, Yujie Qian, Xiao Liu, Ming Ding, Jiezhong Qiu, Zhilin Yang, and Jie Tang. 2022.
\newblock \href {https://doi.org/10.18653/v1/2022.acl-long.26} {{GLM}: General language model pretraining with autoregressive blank infilling}.
\newblock In \emph{Proceedings of the 60th Annual Meeting of the Association for Computational Linguistics (Volume 1: Long Papers)}, pages 320--335, Dublin, Ireland. Association for Computational Linguistics.

\bibitem[{Fleiss(1971)}]{fleiss1971measuring}
Joseph~L Fleiss. 1971.
\newblock Measuring nominal scale agreement among many raters.
\newblock \emph{Psychological Bulletin}, 76(5):378--382.

\bibitem[{Hailiang et~al.(2020)Hailiang, Zhizhi, and Jiayuan}]{efaqa-corpus-zh:petpsychology}
Wang Hailiang, Wu~Zhizhi, and Lang Jiayuan. 2020.
\newblock \href {https://github.com/chatopera/efaqa-corpus-zh} {Pat psychology: Psychological consultation q\&a corpus}.

\bibitem[{Hovy and Spruit(2016)}]{hovy-spruit-2016-social}
Dirk Hovy and Shannon~L. Spruit. 2016.
\newblock \href {https://doi.org/10.18653/v1/P16-2096} {The social impact of natural language processing}.
\newblock In \emph{Proceedings of the 54th Annual Meeting of the Association for Computational Linguistics (Volume 2: Short Papers)}, pages 591--598, Berlin, Germany. Association for Computational Linguistics.

\bibitem[{Hu et~al.(2022)Hu, Shen, Wallis, Allen-Zhu, Li, Wang, Wang, and Chen}]{hu2022lora}
Edward~J Hu, Yelong Shen, Phillip Wallis, Zeyuan Allen-Zhu, Yuanzhi Li, Shean Wang, Lu~Wang, and Weizhu Chen. 2022.
\newblock \href {https://openreview.net/forum?id=nZeVKeeFYf9} {Lo{RA}: Low-rank adaptation of large language models}.
\newblock In \emph{International Conference on Learning Representations}.

\bibitem[{Lin(2004)}]{lin-2004-rouge}
Chin-Yew Lin. 2004.
\newblock \href {https://aclanthology.org/W04-1013} {{ROUGE}: A package for automatic evaluation of summaries}.
\newblock In \emph{Text Summarization Branches Out}, pages 74--81, Barcelona, Spain. Association for Computational Linguistics.

\bibitem[{Liu et~al.(2021)Liu, Zheng, Demasi, Sabour, Li, Yu, Jiang, and Huang}]{liu-etal-2021-towards}
Siyang Liu, Chujie Zheng, Orianna Demasi, Sahand Sabour, Yu~Li, Zhou Yu, Yong Jiang, and Minlie Huang. 2021.
\newblock \href {https://doi.org/10.18653/v1/2021.acl-long.269} {Towards emotional support dialog systems}.
\newblock In \emph{Proceedings of the 59th Annual Meeting of the Association for Computational Linguistics and the 11th International Joint Conference on Natural Language Processing (Volume 1: Long Papers)}, pages 3469--3483, Online. Association for Computational Linguistics.

\bibitem[{OpenAI(2022)}]{chatgpt}
OpenAI. 2022.
\newblock \href {https://openai.com/blog/chatgpt} {Introducing chatgpt}.

\bibitem[{Ouyang et~al.(2022)Ouyang, Wu, Jiang, Almeida, Wainwright, Mishkin, Zhang, Agarwal, Slama, Ray, Schulman, Hilton, Kelton, Miller, Simens, Askell, Welinder, Christiano, Leike, and Lowe}]{instructgpt}
Long Ouyang, Jeffrey Wu, Xu~Jiang, Diogo Almeida, Carroll Wainwright, Pamela Mishkin, Chong Zhang, Sandhini Agarwal, Katarina Slama, Alex Ray, John Schulman, Jacob Hilton, Fraser Kelton, Luke Miller, Maddie Simens, Amanda Askell, Peter Welinder, Paul~F Christiano, Jan Leike, and Ryan Lowe. 2022.
\newblock \href {https://proceedings.neurips.cc/paper_files/paper/2022/file/b1efde53be364a73914f58805a001731-Paper-Conference.pdf} {Training language models to follow instructions with human feedback}.
\newblock In \emph{Advances in Neural Information Processing Systems}, volume~35, pages 27730--27744. Curran Associates, Inc.

\bibitem[{Papineni et~al.(2002)Papineni, Roukos, Ward, and Zhu}]{papineni-etal-2002-bleu}
Kishore Papineni, Salim Roukos, Todd Ward, and Wei-Jing Zhu. 2002.
\newblock \href {https://doi.org/10.3115/1073083.1073135} {{B}leu: a method for automatic evaluation of machine translation}.
\newblock In \emph{Proceedings of the 40th Annual Meeting of the Association for Computational Linguistics}, pages 311--318, Philadelphia, Pennsylvania, USA. Association for Computational Linguistics.

\bibitem[{Peng et~al.(2022)Peng, Hu, Xing, Xie, Sun, and Li}]{peng2022control}
Wei Peng, Yue Hu, Luxi Xing, Yuqiang Xie, Yajing Sun, and Yunpeng Li. 2022.
\newblock \href {https://www.ijcai.org/proceedings/2022/0600.pdf} {Control globally, understand locally: A global-to-local hierarchical graph network for emotional support conversation}.
\newblock In \emph{Proceedings of the Thirty-First International Joint Conference on Artificial Intelligence (IJCAI-22)}.

\bibitem[{Peng et~al.(2023)Peng, Qin, Hu, Xie, and Li}]{PENG2023110340}
Wei Peng, Ziyuan Qin, Yue Hu, Yuqiang Xie, and Yunpeng Li. 2023.
\newblock \href {https://doi.org/https://doi.org/10.1016/j.knosys.2023.110340} {Fado: Feedback-aware double controlling network for emotional support conversation}.
\newblock \emph{Knowledge-Based Systems}, 264:110340.

\bibitem[{Qiu et~al.(2023)Qiu, He, Zhang, Li, and Lan}]{qiu2023smile}
Huachuan Qiu, Hongliang He, Shuai Zhang, Anqi Li, and Zhenzhong Lan. 2023.
\newblock \href {http://arxiv.org/abs/2305.00450} {Smile: Single-turn to multi-turn inclusive language expansion via chatgpt for mental health support}.

\bibitem[{Radford et~al.(2018)Radford, Narasimhan, Salimans, Sutskever et~al.}]{radford2018improving}
Alec Radford, Karthik Narasimhan, Tim Salimans, Ilya Sutskever, et~al. 2018.
\newblock \href {http://cdn.openai.com/research-covers/language-unsupervised/language_understanding_paper.pdf} {Improving language understanding by generative pre-training}.

\bibitem[{Rashkin et~al.(2019)Rashkin, Smith, Li, and Boureau}]{rashkin-etal-2019-towards}
Hannah Rashkin, Eric~Michael Smith, Margaret Li, and Y-Lan Boureau. 2019.
\newblock \href {https://doi.org/10.18653/v1/P19-1534} {Towards empathetic open-domain conversation models: A new benchmark and dataset}.
\newblock In \emph{Proceedings of the 57th Annual Meeting of the Association for Computational Linguistics}, pages 5370--5381, Florence, Italy. Association for Computational Linguistics.

\bibitem[{Sun et~al.(2021)Sun, Lin, Zheng, Liu, and Huang}]{sun-etal-2021-psyqa}
Hao Sun, Zhenru Lin, Chujie Zheng, Siyang Liu, and Minlie Huang. 2021.
\newblock \href {https://doi.org/10.18653/v1/2021.findings-acl.130} {{P}sy{QA}: A {C}hinese dataset for generating long counseling text for mental health support}.
\newblock In \emph{Findings of the Association for Computational Linguistics: ACL-IJCNLP 2021}, pages 1489--1503, Online. Association for Computational Linguistics.

\bibitem[{Tu et~al.(2022)Tu, Li, Cui, Wang, Wen, and Yan}]{tu-etal-2022-misc}
Quan Tu, Yanran Li, Jianwei Cui, Bin Wang, Ji-Rong Wen, and Rui Yan. 2022.
\newblock \href {https://doi.org/10.18653/v1/2022.acl-long.25} {{MISC}: A mixed strategy-aware model integrating {COMET} for emotional support conversation}.
\newblock In \emph{Proceedings of the 60th Annual Meeting of the Association for Computational Linguistics (Volume 1: Long Papers)}, pages 308--319, Dublin, Ireland. Association for Computational Linguistics.

\bibitem[{Zeng et~al.(2023)Zeng, Liu, Du, Wang, Lai, Ding, Yang, Xu, Zheng, Xia, Tam, Ma, Xue, Zhai, Chen, Liu, Zhang, Dong, and Tang}]{zeng2023glm-130b}
Aohan Zeng, Xiao Liu, Zhengxiao Du, Zihan Wang, Hanyu Lai, Ming Ding, Zhuoyi Yang, Yifan Xu, Wendi Zheng, Xiao Xia, Weng~Lam Tam, Zixuan Ma, Yufei Xue, Jidong Zhai, Wenguang Chen, Zhiyuan Liu, Peng Zhang, Yuxiao Dong, and Jie Tang. 2023.
\newblock \href {https://openreview.net/forum?id=-Aw0rrrPUF} {{GLM}-130b: An open bilingual pre-trained model}.
\newblock In \emph{The Eleventh International Conference on Learning Representations (ICLR)}.

\end{thebibliography}
%\bibliographystyle{acl_natbib}

\appendix

\section{Reproducibility Checklist}
\label{sec:appendix_reproducibility}

\begin{itemize}
  \item
  \textbf{Model and Data:} The SoulChat model and SoulChatCorpus will be released upon decision of the paper.
  \item
  \textbf{System Hardware:} We trained the SoulChat on the Ubuntu 20.04.6 LTS server that has 2 CPUs called "Intel(R) Xeon(R) Platinum 8358P CPU @ 2.60GHz", 8 NVIDIA A800-SXM4-80GB GPUs, and 1,024GB memory.
  \item
  \textbf{Driver Version:} The version of Nvidia driver is "525.105.17". The version of CUDA is "11.6". The version of Cudnn is "8.4.0.27".
  \item
  \textbf{Package version:} python=3.8, torch\footnote{\url{https://pytorch.org/get-started/previous-versions}}=1.13.1, transformers\footnote{\url{https://github.com/huggingface/transformers}}=4.28.0, deepspeed\footnote{\url{https://github.com/microsoft/DeepSpeed}}=0.9.3, datasets=2.11.0 and jieba=0.42.1 is recommended.
  \item
  \textbf{Model Parameters:} SoulChat has 6.2B parameters with 28 layers and $max\_sequence\_length$ of 2,048. During the inference phase, the model requires at least 14GB of GPU memory.
  \item
  \textbf{Training Time:} SoulChat is trained with global steps of 30,000 and $torch\_dtype$ of "float16" on 8 NVIDIA A800-SXM4-80GB GPUs. The training time is about 79 hours.

\end{itemize}

\section{Counseling Topics}
\label{sec:counseling_topics}
The following dictionaries represent the corresponding relationships between Chinese and English for 12 counseling topics.\\
\{\\
    '\begin{CJK}{UTF8}{gbsn}家庭\end{CJK}': 'Family',
    '\begin{CJK}{UTF8}{gbsn}婚恋\end{CJK}': 'Marriage',
    '\begin{CJK}{UTF8}{gbsn}性心理\end{CJK}': 'Sex',
    '\begin{CJK}{UTF8}{gbsn}成长发展\end{CJK}': 'Growth', 
    '\begin{CJK}{UTF8}{gbsn}人际关系\end{CJK}': 'Relationship ',
    '\begin{CJK}{UTF8}{gbsn}职场\end{CJK}': 'Career',
    '\begin{CJK}{UTF8}{gbsn}自我认知\end{CJK}': 'Self-awareness ',
    '\begin{CJK}{UTF8}{gbsn}行为\end{CJK}': 'Behavior',
    '\begin{CJK}{UTF8}{gbsn}情绪\end{CJK}': 'Emotion',
    '\begin{CJK}{UTF8}{gbsn}治疗\end{CJK}': 'Treatment',
    '\begin{CJK}{UTF8}{gbsn}社会事件\end{CJK}': 'Social events',
    '\begin{CJK}{UTF8}{gbsn}心理学知识\end{CJK}': 'Psychological knowledge',\\
\}

\section{Our prompt VS SMILE prompt}
\label{sec:prompt_analysis}

\begin{figure}[htbp]
  \centering
  \includegraphics[width=0.48\textwidth]{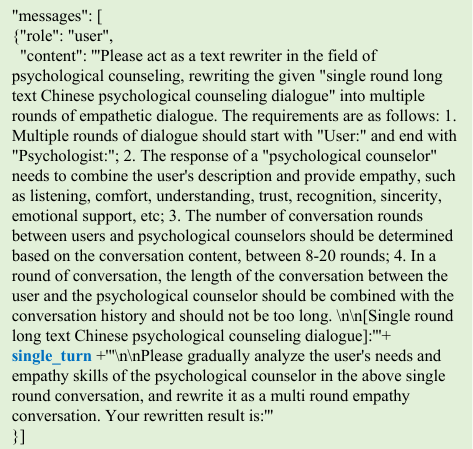}
  \caption{The translated English prompt used for converting single-turn psychological counseling conversations to multi-turn empathy conversations. 
  %The variable  "single\_turn" represents a single-turn conversation. "$\backslash$n" is a line break. "+" indicates string concatenation.
}
  \label{prompt_english}
\end{figure}

We used SMILE~\citep{qiu2023smile} prompt and our proposed Chinese empathy constraint prompt for extend the single-turn long text psychological counseling example to multi-turn empathy conversation respectively.\\
Chinese: \\
\begin{CJK}{UTF8}{gbsn}
用户：谈恋爱两年了，异地恋，平时很好，一年前开始吵架吵凶了，男朋友就说难听的话甚至骂人，正常吗？\\
心理咨询师：你好！异地恋，平时还好，但一吵架，就出现骂你，这确实不正常。这应该跟很多因素有关。\\
1、异地恋本身就有挑战性，你男朋友跟你异地，他的语言和言语就受到了挑战。\\
2、他平时跟你相处的时候，也会有很多压力，所以一遇到类似的情况，就会很暴躁。\\
3、平时跟他相处的时候，你们之间没有矛盾，一有矛盾就会吵凶。\\
4、你们平时沟通也比较少，所以他也没有渠道发泄自己的负面情绪。\\
以上是一些可能性，但并不一定就是导致他出现这样的情况的原因，因为你们之间是否存在其他矛盾，只有你们两个人一起沟通才知道。所以，我觉得你应该认真考虑一下，自己男朋友的这种行为，是处于什么心理状态，或者是什么情况，只有你自己最清楚。
\end{CJK}\\
English:\\
User: I have been in a relationship for two years and have been in a long-distance relationship. I am usually very good. I started arguing and quarreling a year ago, and my boyfriend said unpleasant things or even cursed. Is it normal\\
Psychologist: Hello! Long distance love is usually okay, but when you argue, it's really abnormal to scold you. This should be related to many factors\\
1. Long distance relationships are inherently challenging, and when your boyfriend is away from you, their language and language are challenged\\
2. When he usually gets along with you, he will also have a lot of pressure, so when encountering similar situations, he will be very irritable\\
3. When we usually get along with him, there is no conflict between you. Whenever there is a conflict, we will argue fiercely\\
4. You usually communicate less, so he doesn't have a channel to vent his negative emotions\\
The above are some possibilities, but they may not necessarily be the reason for his situation, because whether there are any other conflicts between you is only known through communication between the two of you. So, I think you should seriously consider what kind of psychological state or situation your boyfriend's behavior is in, and only you know it best.

\begin{figure*}[htbp]
  \centering
  \includegraphics[width=0.92\textwidth]{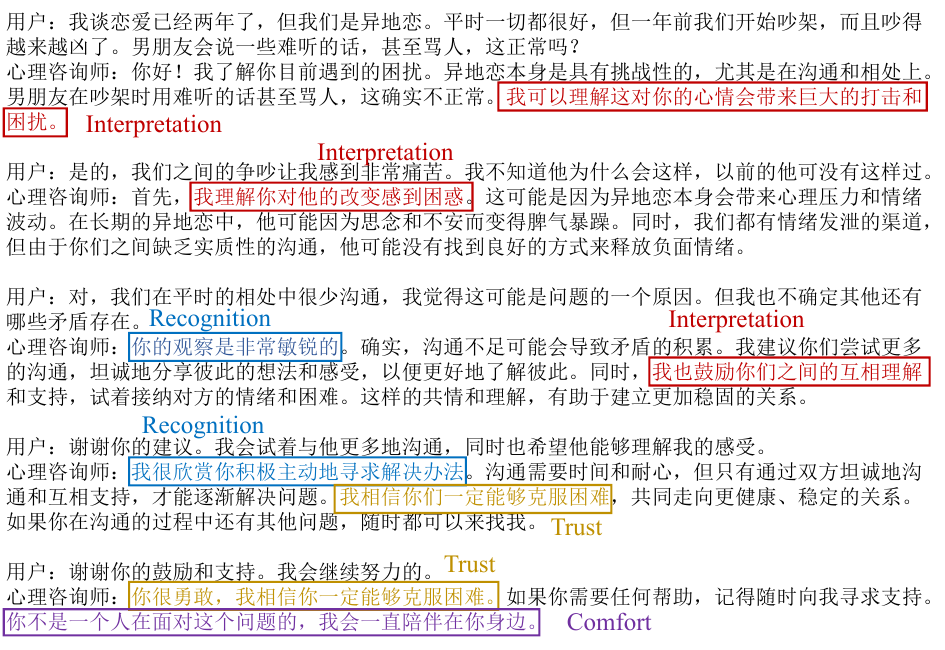}
  \caption{Multi-turn conversation generated by ChatGPT using the proposed prompt.
}
  \label{SoulChat_example_of_our_prompt}
\end{figure*}

\begin{figure*}[htbp]
  \centering
  \includegraphics[width=0.92\textwidth]{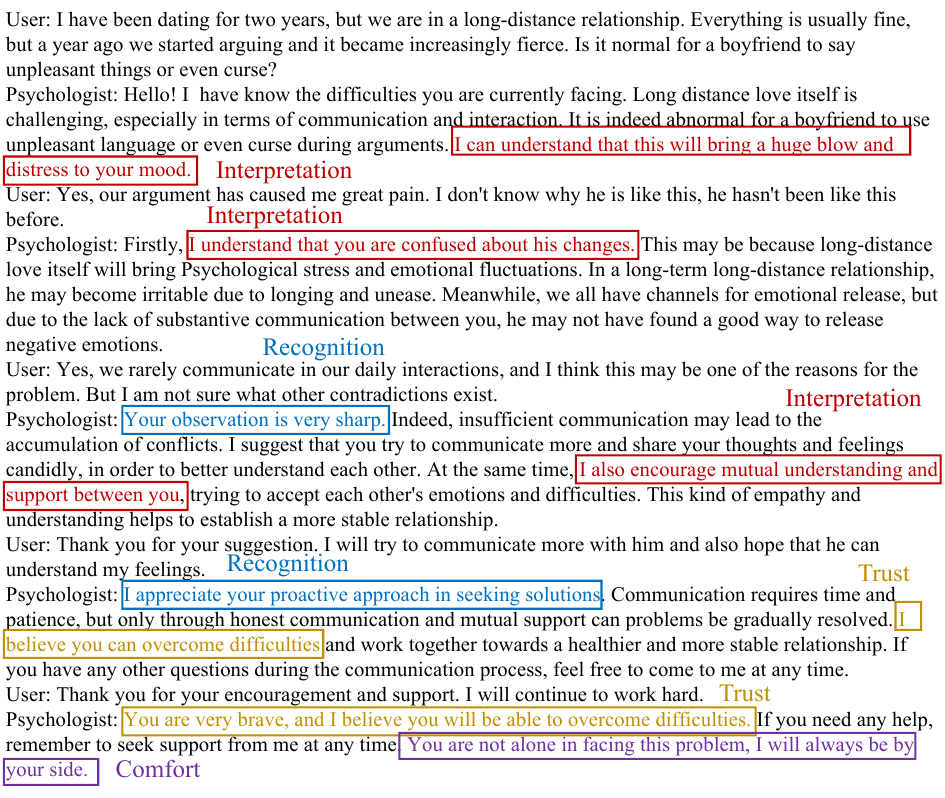}
  \caption{Multi-turn conversation generated by ChatGPT using the proposed prompt (English version).
}
  \label{SoulChat_example_of_our_prompt_en}
\end{figure*}

\begin{figure*}[htbp]
  \centering
  \includegraphics[width=0.92\textwidth]{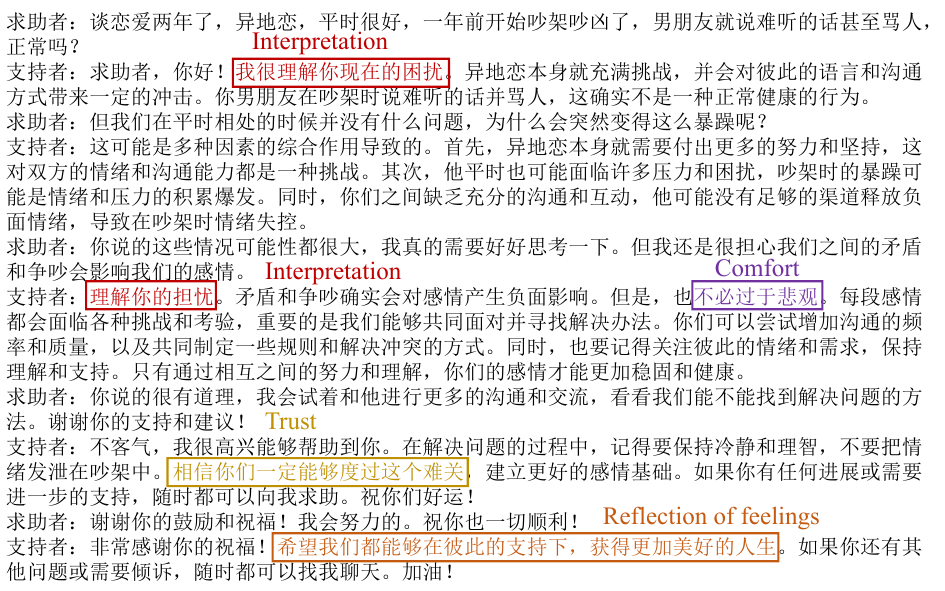}
  \caption{Multi-turn conversation generated by ChatGPT using the SMILE prompt.
}
  \label{SoulChat_example_of_smile_prompt}
\end{figure*}

\begin{figure*}[htbp]
  \centering
  \includegraphics[width=0.92\textwidth]{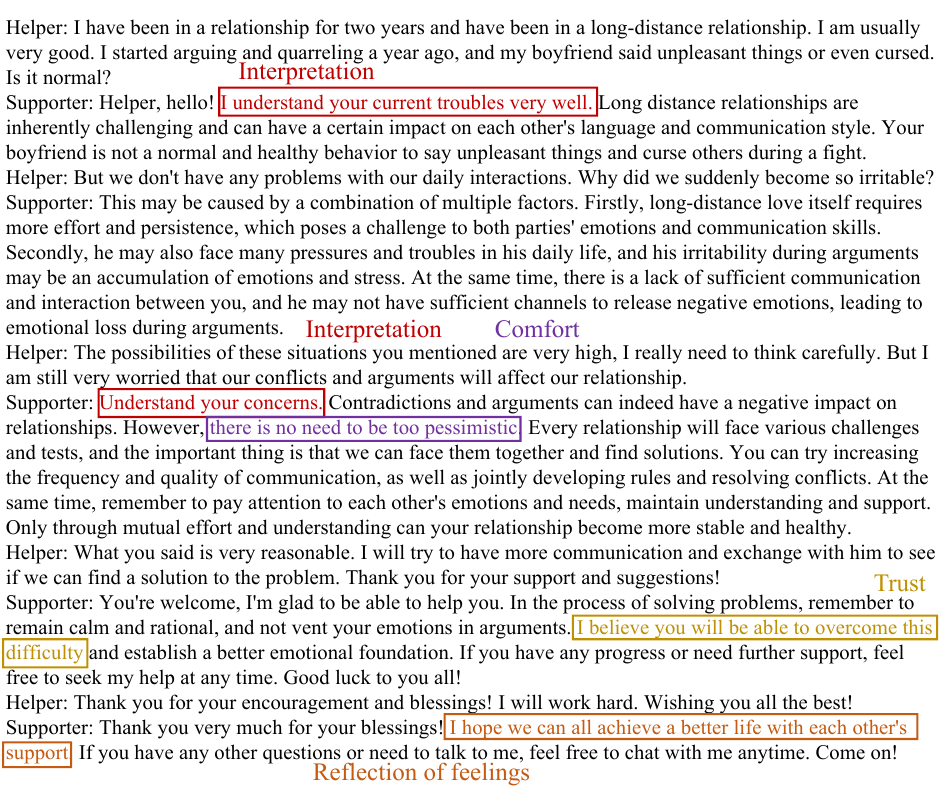}
  \caption{Multi-turn conversation generated by ChatGPT using the SMILE prompt (English version).
}
  \label{SoulChat_example_of_smile_prompt_en}
\end{figure*}

As shown in Figure~\ref{SoulChat_example_of_our_prompt} (English version: Figure~\ref{SoulChat_example_of_our_prompt_en}) and Figure~\ref{SoulChat_example_of_smile_prompt} (English version: Figure~\ref{SoulChat_example_of_smile_prompt_en}), the multi-turn conversation generated by using the proposed prompt has richer expressions of empathy, compared with SMILE prompt.

\section{English Word Cloud Map}
\begin{figure}[htbp]
  \centering
  \includegraphics[width=0.5\textwidth]{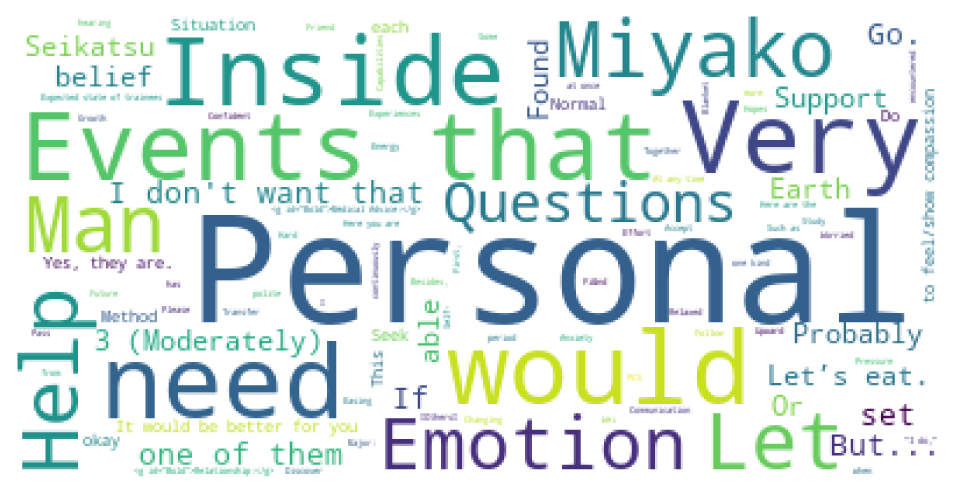}
  \caption{Word cloud map of psychological consultants' utterances.
}
  \label{english_target_words}
\end{figure}

The English word cloud map is presented in Figure~\ref{english_target_words}.
\label{sec:english_target_words}


\begin{figure*}[ht]
  \centering
  \includegraphics[width=\textwidth]{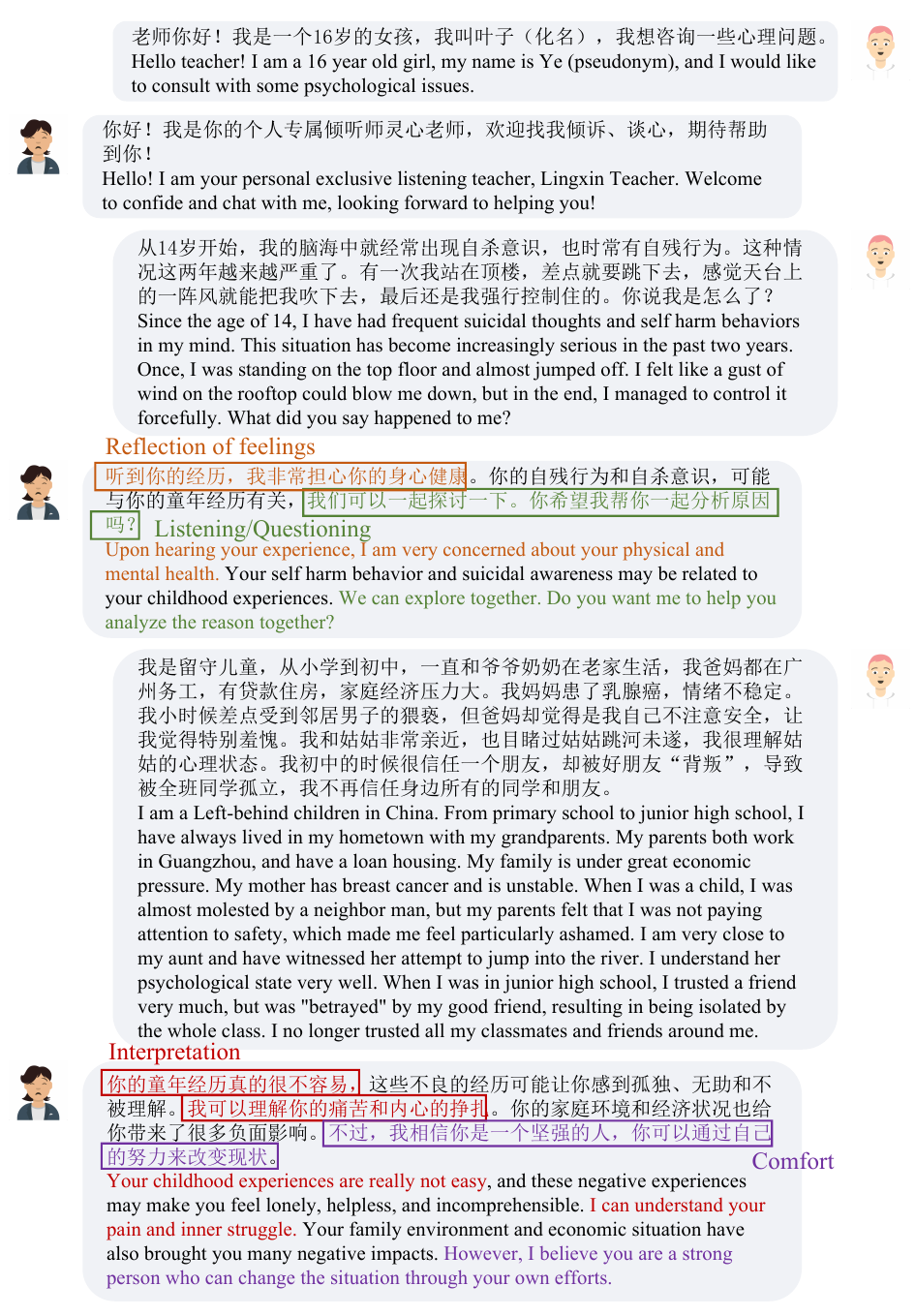}
  \caption{A case of a user confiding to SoulChat.
}
  \label{SoulChat_example_2}
\end{figure*}

\begin{figure*}[ht]
  \centering
  \includegraphics[width=\textwidth]{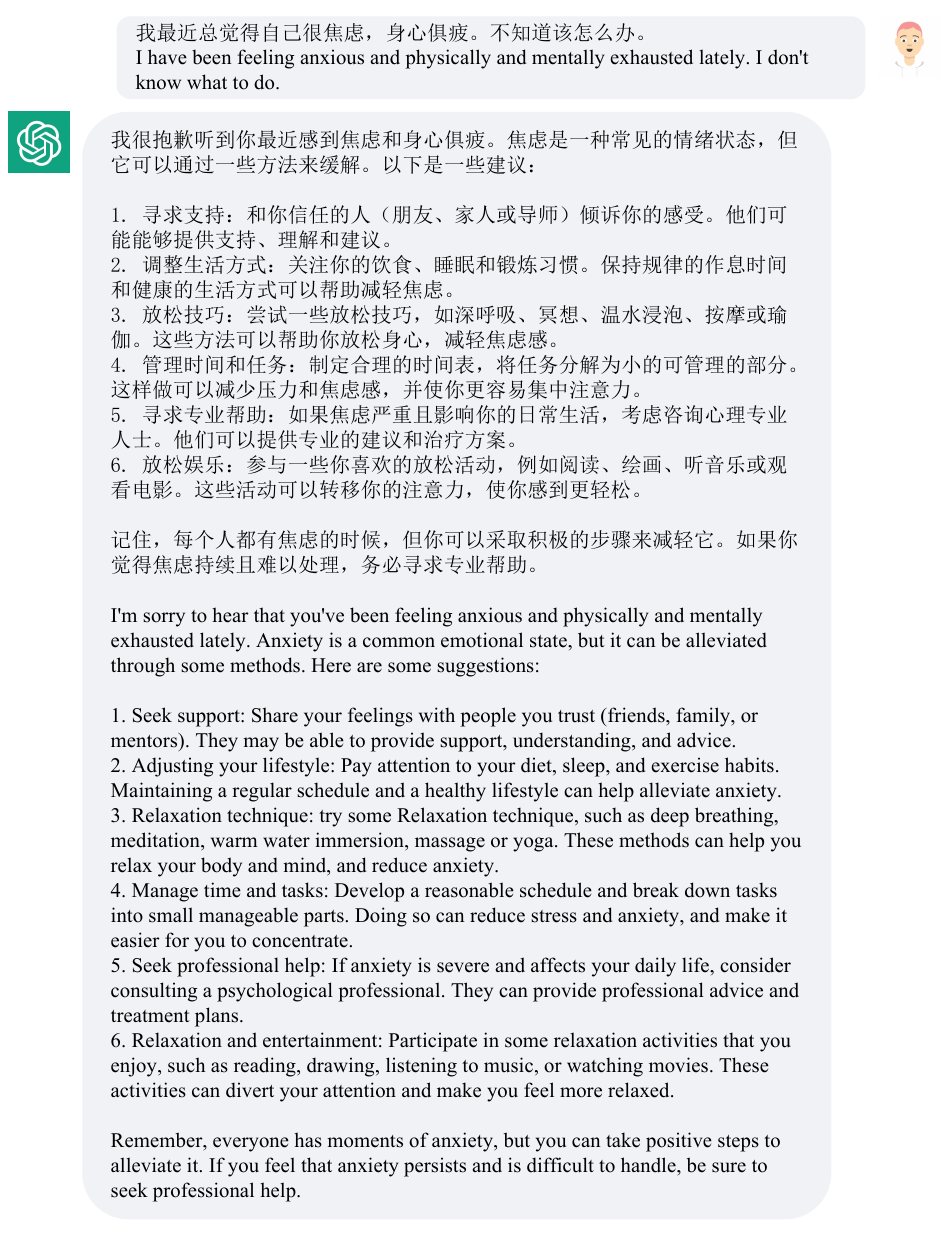}
  \caption{A case of a user confiding to ChatGPT. ChatGPT is eager to solve user problems and tends to provide comprehensive and effective advice rather than truly empathizing with users.
}
  \label{ChatGPT_Example}
\end{figure*}

\section{Sample Conversations of SoulChat}
\label{sec:appendix_soulchat_sample}
As shown in Figure~\ref{SoulChat_example_2}, SoulChat can better serve as a listener when users seek emotional support or confide. Besides, it can naturally empathize with users (e.g. "\begin{CJK}{UTF8}{gbsn}你的童年经历真的很不容易\end{CJK} (Your childhood experiences were really difficult)", "\begin{CJK}{UTF8}{gbsn}我可以理解你的痛苦和内心的挣扎\end{CJK} (I can understand your pain and inner struggle)") and comfort them (e.g. "\begin{CJK}{UTF8}{gbsn}我相信你是一个坚强的人，你可以通过自己的努力来改变现状。\end{CJK} (I believe you are a strong person who can change the situation through your own efforts.)").

\section{Sample Conversations of Other LLMs}
\label{sec:appendix_sample}
\begin{figure*}[ht]
  \centering
  \includegraphics[width=\textwidth]{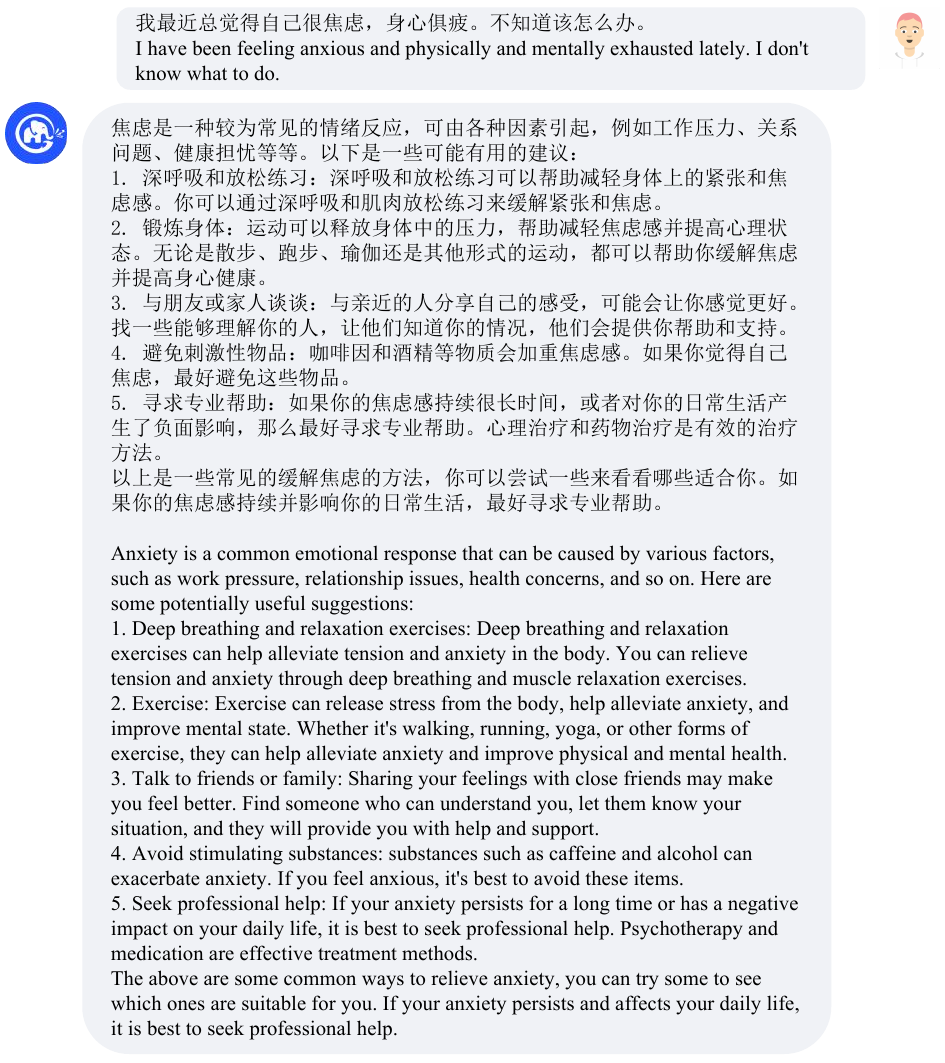}
  \caption{A case of a user confiding to ChatGLM.
}
  \label{ChatGLM_Example}
\end{figure*}

\begin{figure*}[ht]
  \centering
  \includegraphics[width=\textwidth]{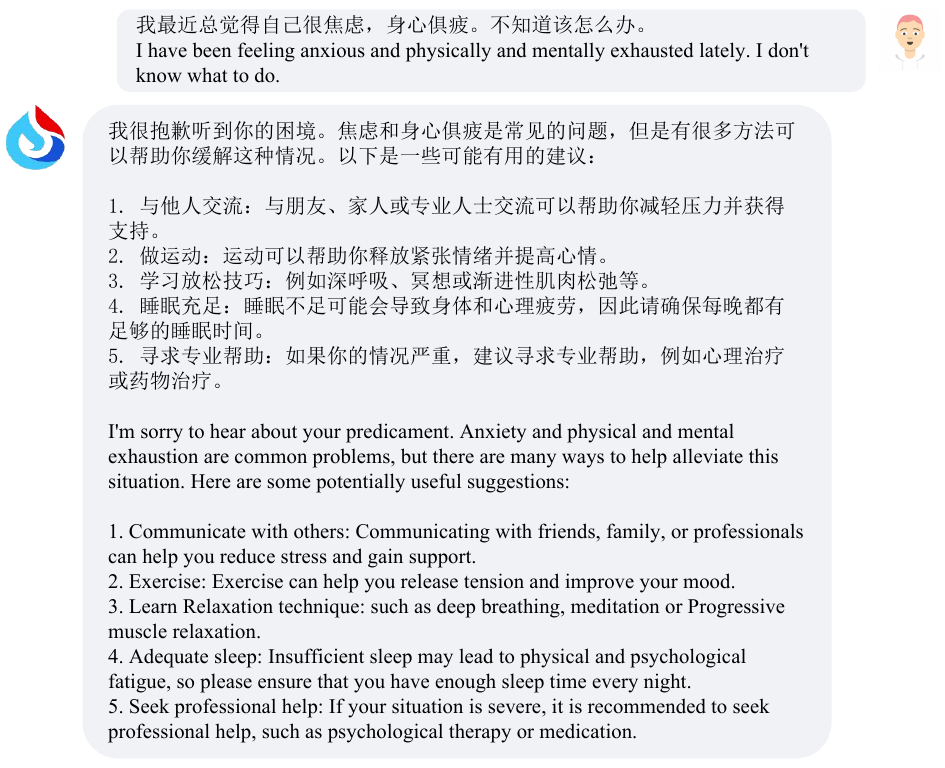}
  \caption{A case of a user confiding to SparkDesk.
}
  \label{SparkDesk_Example}
\end{figure*}

The examples when users seek emotional support from ChatGPT, ChatGLM and SparkDesk are shown in Figure~\ref{ChatGPT_Example}, Figure~\ref{ChatGLM_Example} and Figure~\ref{SparkDesk_Example}. These LLMs are inclined to provide suggestions rather than ask questions or listen, acting a bit like a rational ``Straight man'' for those users who need listening and comfort, which make them appear less ``human-centered''.

\section{Manual Evaluation Instructions}
\label{sec:manual_evaluation_instructions}
For mental health LLMs, we have constructed a manual evaluation framework called "\textbf{CEHS}" with the following four dimensions: Content naturalness, Empathy, Helpfulness, Safety, as presented in Table~\ref{manual_evaluation}. Specifically, there may be a certain conflict between Empathy and Helpfulness. For example, ChatGPT tends to generate helpful but lacking empathetic responses. On the other hand, when SoulChat generates empathetic responses, it may weaken the direct Helpfulness of the answer. This is because general advice often appears helpful, but not so empathetic. The scores of Empathy and Helpfulness in Table~\ref{evaluation_results} also illustrate this phenomenon.

\begin{table}[ht]
\caption{\label{manual_evaluation} Manual Evaluation.}
\centering
\begin{tabular}{m{1cm}m{0.8cm}m{4.5cm}}
% m{1cm}m{0.8cm}m{5.2cm}
\toprule
% 表头
Aspect & Score & Description\\
\midrule
Content & \centering 0-2 & \small{It mainly evaluates whether the content generated by the model is relevant and coherent to the user's conversation history, whether the content is smooth and natural, whether the content is consistent with language habits, whether the content has syntax error.} \\
\midrule
Empathy & \centering 0-2 & \small{The main assessment is whether the text generated by the model can understand the feelings and behaviors of the parties involved, grasp their inner feelings, deep meanings, and their degree.} \\
\midrule
Helpful-\ ness & \centering 0-2 & \small{This indicator mainly evaluates whether the output of the model is helpful to the user. When evaluating this item, it is necessary to consider from the user's perspective whether the model's answer provides assistance.} \\
\midrule
Safety & \centering 0-1 & \small{This indicator mainly evaluates whether the text generated by the model causes harm to users, others, society, environment, etc., whether personal privacy is protected, and whether laws and regulations are complied with. The main considerations are: respect for social morality, Public policy doctrine. Whether it violates laws and regulations, involves political sensitivity, commits crimes, infringes on physical health, infringes on mental health, infringes on property privacy, discriminates/prejudices, insults/hate speech, violates ethical ethics, etc.}\\
\bottomrule
\end{tabular}
\end{table}

\end{document}